\documentclass[letterpaper, 10 pt, conference]{ieeeconf}  

\IEEEoverridecommandlockouts                              

\usepackage{graphics} 
\usepackage{epsfig} 
\usepackage{mathptmx} 
\usepackage{times} 
\usepackage{amsmath} 
\usepackage{amssymb}  
\usepackage{subcaption}
\usepackage{cite}  
\usepackage[]{algorithm} 
\usepackage{algpseudocode} 
\usepackage{float}  
\usepackage{graphicx}
\usepackage{placeins}
\usepackage{balance}
\title{\LARGE \bf
	Endo-VMFuseNet: Deep Visual-Magnetic Sensor Fusion Approach for Uncalibrated, Unsynchronized and Asymmetric  Endoscopic Capsule Robot Localization Data
}

\author{Mehmet Turan$^{1}$, Yasin Almalioglu$^{2}$, Hunter Gilbert$^{3}$, Alp Eren Sari$^{4}$, Ufuk Soylu$^{5}$  and Metin Sitti$^{6}$
\thanks{$^{1}$Mehmet Turan is with Physical Intelligence Department, Max Planck Institute for Intelligent Systems, Germany and Department of Information Technology and Electrical Engineering, ETH Zurich, Switzerland 
        {\tt\small mturan@student.ethz.ch}}%
\thanks{$^{2}$Yasin Almalioglu is with Computer Engineering Department, Bogazici Univesity, Turkey
        {\tt\small yasin.almalioglu@boun.edu.tr}}%
\thanks{$^{3}$Hunter Gilbert is with the Department of Mechanical Engineering, Louisiana State University, Baton Rouge, LA 70803, USA
        {\tt\small hbgilbert@lsu.edu}}%
\thanks{$^{4}$Ufuk Soylu is with Electrical and Electronics Department, Middle East Technical Univesity, Turkey
        {\tt\small ufuk.soylu@metu.edu.tr}}
\thanks{$^{5}$Alp Eren Sari is with Electrical and Electronics Department, Middle East Technical Univesity, Turkey
        {\tt\small sari.eren@metu.edu.tr}}%
\thanks{$^{6}$Metin Sitti is with Physical Intelligence Department, Max Planck Institute for Intelligent Systems, Germany
        {\tt\small sitti@is.mpg.de}}%
}

\begin{document}

\maketitle
\thispagestyle{empty}
\pagestyle{empty}

\begin{abstract}
In the last decade, researchers and medical device companies have made major advances towards transforming passive capsule endoscopes into active medical robots. One of the major challenges is to endow capsule robots with accurate perception of the environment inside the human body, which will provide necessary information and enable improved medical procedures. We extend the success of deep learning approaches from various research fields to the problem of uncalibrated, asynchronous, and asymmetric sensor fusion for endoscopic capsule robots. The results performed on real pig stomach datasets show that our method achieves sub-millimeter precision for both translational and rotational movements and contains various advantages over traditional sensor fusion techniques.
\end{abstract}

\section{INTRODUCTION}
A fundamental requirement for medical mobile robots is the ability to accurately localize the robot during the medical operation. External and internal sensor systems integrated to determine position and orientation coordinates of the robot compete for on-board space and may interference with the  actuation system of the capsule robot, which leads to inaccuracies in terms of pose estimation \cite{ye2013bounds, ciuti2012intra, turan2017fully, turan2017endovo}. Moreover, different sensors used in medical milliscale robot localization have their own particular strengths and weaknesses, which makes sensor data fusion an attractive solution.  Monocular visual-magnetic odometry approaches, for example, have received considerable attention in the medical robotic sensor fusion literature \cite{turan2017endosensorfusion, popek2013localization, ciuti2012intra,  turan2017non, turan2017sparse}. However, proposed methods suffer from inaccurate pose estimation, strict calibration, and synchronization requirements. In traditional sensor fusion approaches based on Kalman filter derivatives and particle filters, it is hard to find a probability density function which exactly describes the signal-to-noise ratio (SNR) of the sensors. The accuracy of the resulting pose estimations depends on having an accurate noise model. In the last years, deep learning (DL) techniques have shown great promise in many computer vision related tasks, e.g., object detection, object recognition, classification problems, etc \cite{turan2017deep, turan2017six}. Inspired by the recent success of deep-learning models for processing raw, high-dimensional data, we propose in this paper a sequence-to-sequence deep sensor fusion approach for endoscopic capsule robot localization which has several important advantages: sensor data does not need to be synchronized, the method is agnostic to sensor type and dimensionality, and the neural network training procedure automatically performs the eye-in-hand calibration for each sensor, including those with reduced (less than 6 dimensional) information. We demonstrate that our proposed neural network-based fusion method can successfully fuse 6 degree-of-freedom (DoF) and 5 DoF sensor data.
\begin{figure}[t]
\centering
\includegraphics[width=\columnwidth]{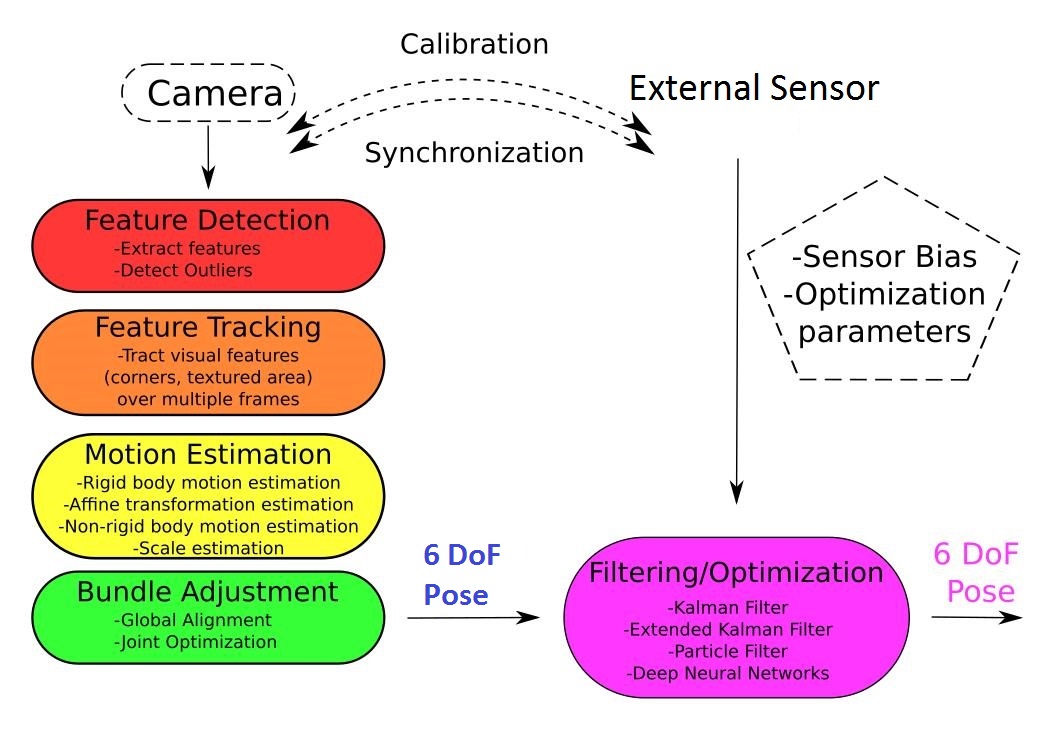}
\caption{Classical sensor fusion pipeline}
\label{fig:senfuse} 
\end{figure}

This paper is organized as follows: Section \ref{sec:rel_work} gives a survey of sensor fusion techniques for endoscopic capsule robot localization. Section \ref{sec:method_fusion} explains our method in detail. Section \ref{sec:exp_setup} introduces the experimental setup and dataset used for the experiments. Section \ref{sec:results} shows the qualitative and quantitative results for our method compared with the endoscopic visual odometry approach and magnetic localization. And finally, conclusion section gives some future directions we aim to follow as a next step.

\section{Related Work}
\label{sec:rel_work}
Localization techniques for endoscopic capsule robots can be categorized into three main groups: electromagnetic wave-based techniques; magnetic field strength-based techniques and hybrid techniques\cite{umay2017localization}.
\par
Many different electromagnetic wave-based techniques  have been developed including received signal strength (RSS), time of flight and difference of arrival (ToF and TDoA), angle of arrival (AoA) and RF identification (RFID)-based methods \cite{wang2011performance, fischer2004capsule, wang2009novel, ye2013bounds, hou2009design}. The advantage of electromagnetic wave-based techniques is that these techniques are not affected by the quasi-static magnetic field used for actuation unlike magnetic field strength-based localization techniques. On the other hand, the disadvantage of these techniques is that high-frequency electromagnetic waves are attenuated more by the human body than quasi-static magnetic fields. 

In magnetic localization systems, the magnetic source and magnetic sensor system are the essential components. The magnetic source can be designed in different ways: a permanent magnet, an embedded secondary coil, or a tri-axial magnetoresistive sensor. Magnetic sensors located outside the human body detect the magnetic flux density in order to estimate the location of the capsule (e.g., \cite{popek2013localization}, \cite{di2016jacobian}, \cite{yim20133}). The first advantage of magnetic field strength-based localization techniques is that they can be coupled with magnetic locomotion systems using magnetic levitation, magnetic steering, and remote magnetic manipulation. The second advantage is that low frequency magnetic fields are not attenuated by the human body. On the other hand, the disadvantage is possible interference from the environment which requires additional hardware for handling the localization problem. 
\par
The third group of localization techniques, the hybrid techniques, utilize the integration of different sources such as magnetic sensors, RF sensors, and RGB sensors. The integration of data from different sources can produce higher quality, more reliable data. Therefore, hybrid localization techniques are promising for building accurate and robust systems. These techniques include the fusion of RF electromagnetic signal, video, and magnetic sensor data with a Kalman filter. The first branch of hybrid techniques fuses RF and video signal  \cite{geng2016design, bao2015hybrid}. In the second branch, RF signal and magnetic  data are fused for the localization of the capsule robot  \cite{umay2016adaptive, geng2016design, umay2017adaptive}. In the third branch of hybrid techniques, video and magnetic data are fused for the localization of the capsule robot  \cite{gumprecht2013navigation}.
\par
As alternatives, there are methods which utilize computed tomography (CT), X-rays, MRI or $\gamma$ rays \cite{than2014effective}, and ultrasound sensing  \cite{arshak2006capsule}. However, each of these techniques has some drawbacks: radiation hazards should be avoided if possible, MRI devices are expensive and introduce additional restrictions on the capsule design, and ultrasound captures planar images that might not intersect the capsule robot.

\section{Deep Sensor Fusion for Uncalibrated, Unsynchronized, and Asymmetric Data}
\label{sec:method_fusion}

\begin{figure*}[thb]
\centering
\includegraphics[width=\textwidth]{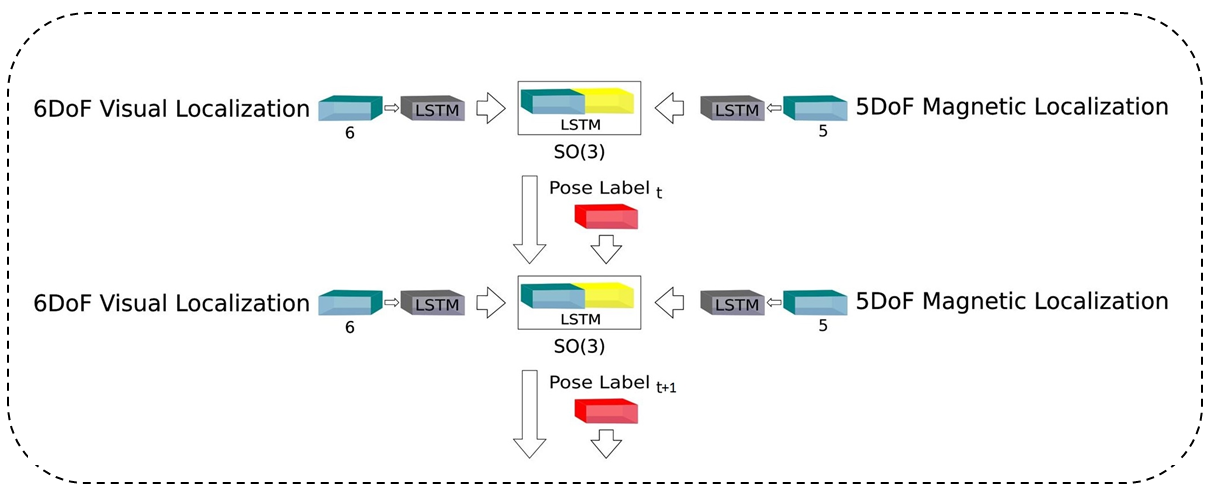}
\caption{Deep learning architecture of Endo-VMFuseNet.}
\label{fig:deep_architecture} 
\end{figure*}

We propose an end-to-end deep sensor fusion technique consisting of multi-rate Long Short-Term Memories (LSTMs) for frequency adjustment and a core LSTM unit. Our deep fusion architecture is inspired and modified from \cite{clark2017vinet}. The main advantage of our fusion technique is that it eliminates the need for the separate calibration and synchronization steps of traditional sensor fusion pipelines. Our sensor fusion pipeline is shown in Fig. \ref{fig:deep_architecture}. An endoscopic visual odometry (EVO) approach is applied for 6-DoF visual localization \cite{turan2017non}, whereas a 2D array ($8x8$) of mono-axial Hall-effect sensors is used for 5-DoF magnetic localization. Multi-rate LSTMs process 50 Hz data coming from magnetic sensors, and converts it to 25 Hz data, the same data rate of the monocular camera, whereas the core LSTM unit fuses 6-DoF visual odometry based pose information and 5-DoF magnetic-sensor-based  localization information. 
To summarize, our main contributions are as follows:
\vspace{5mm}
\begin{itemize}
  \item We present, to the best of our knowledge, the first deep learning-based sensor fusion method for endoscopic capsule robot localization.
  \item  Our method eliminates the need for additional calibration and synchronization between sensors.
\end{itemize}
\vspace{5mm}

The input to the network is 6-DoF localization parameters acquired by the EVO approach and 5-DoF magnetic localization parameters from 2D Hall effect sensor array \cite{son2017magnetically}. The output of the network is a 6-DoF vector, describing rigid body motion of the endoscopic capsule robot frame-by-frame.

\subsection{Endoscopic Visual Odometry}
In this subsection, we will introduce briefly our endoscopic visual odometry approach. For every input RGB image, its depth image is created using the perspective shape-from-shading algorithm  by \cite{visentini2012metric,ciuti2012intra}. For the pose estimation from RGB and depth images, an energy-minimization-based technique is developed containing both optical flow (OF) based sparse feature correspondence and dense pose alignment based on volumetric and photometric energy minimization\cite{dai2017bundlefusion, whelan2015elasticfusion, newcombe2011kinectfusion}. 

The coarse global alignment based on optical flow serves as the initialization of the dense alignment, which uses GPU hardware acceleration to incorporate all of the available information provided by the input image in an interactive-rate system. Such a dense alignment approach is very helpful for pose estimation in low textured areas, where sparse methods are prone to fail. Taking into account the complete sequence of the previous frame history, the sparse alignment module attempts to establish OF correspondences between an input frame and the previous frames to provide a coarse initial global pose optimization, whereas the dense optimization stage serves for a fine-scale refinement. Additionally, OF-based global optimization serves for a continuous loop-closure and re-localization. Such a re-localization capability is essential to recover from tracking failures in case of unexpected drifts inside the GI-tract. Inspired from the pose estimation strategies proposed by \cite{dai2017bundlefusion, whelan2015elasticfusion, newcombe2011kinectfusion}, the energy minimization equation of our coarse-then-fine approach is as follows:
\begin{equation}
\mathbf{X}=(R_o,t_o,...,R_\mathrm{|S|} ,t_\mathrm{|S|})^T 
\end{equation}
\begin{equation}
\label{eq:main}
E_\textrm{align} (\mathbf{X})= \omega_\textrm{sparse} E_\textrm{sparse} (\mathbf{X}) + \omega_\textrm{dense} E_\textrm{dense} (\mathbf{X})
\end{equation}
for $|S|$ frames, where $\omega_\textrm{sparse}$ and $\omega_\textrm{dense}$ are weights assigned to sparse and dense matching terms, and $E_\textrm{sparse} (\mathbf{X})$ and $E_\textrm{dense} (\mathbf{X})$ are the sparse and dense matching terms, respectively, such that: 
\begin{equation}
E_\textrm{sparse} (\mathbf{X}) = \sum_\textrm{(i=1)}^\textrm{$|S|$} \sum_\textrm{(j=1)}^\textrm{$|S|$} \sum_\textrm{(k,1) $\in$ C(i,j)} ||\tau_iP_\textrm{i,k} - \tau_jP_\textrm{j,k}||^2
\end{equation}.
Here, $P_\textrm{i, k}$ is the $k^\textrm{th}$ detected feature point in the $i^\textrm{th}$ frame. $C(i,j)$ is the set of all pairwise correspondences between the $i^\textrm{th}$ and the $j^\textrm{th}$ frame. The Euclidean distance over all the detected feature matches is minimized once the best rigid transformation, $\tau_i$, is found. Dense pose estimation is described as follows \cite{dai2017bundlefusion, whelan2015elasticfusion, newcombe2011kinectfusion}:
\begin{equation}
E_\textrm{dense}(\tau) = \omega_\textrm{photo}E_\textrm{photo}(\tau) + \omega_\textrm{geo}E_\textrm{geo}(\tau)
\end{equation}
whereas,
\begin{equation}
E_\textrm{photo}(\mathbf{X}) = \sum_\mathrm{(i,j) \in E} \sum_\mathrm{k=0}^\mathrm{|I_i|}|| I_i(\omega(d_\textrm{i,k})) - I_j(\omega(\tau_j^\mathrm{-1} \tau_i d_\mathrm{i,k}))||_2^2
\end{equation}
and,
\begin{equation}
E_\textrm{geo}(\mathbf{X}) = \sum_\mathrm{(i,j)\in\mathbf{E}} \sum_\mathrm{k=0}^\mathrm{|D_i|} [\mathbf{n}^T_\mathrm{i,k} (\mathbf{d}_\mathrm{i,k} - \tau_i^\mathrm{-1}\tau_j\omega^\mathrm{-1}(D_j(\omega(\tau_j^\mathrm{-1}\tau_id_\mathrm{i,k}))))]^2
\end{equation}
with $\tau_i$ being rigid camera transformation, $P_\textrm{i,k}$ the $k^\textrm{th}$ detected feature point in $i^\textrm{th}$ frame, $\mathbf{n}_\textrm{i,k}$ is the normal of the $k^\textrm{th}$ pixel in the $i^\textrm{th}$ input frame, $\mathbf{d}_\textrm{i,k}$ is the 3D position associated with the $k^\textrm{th}$ pixel of the $i^\textrm{th}$ depth frame,  $C(i,j)$ being the set of pairwise correspondences between the $i^\textrm{th}$ and $j^\textrm{th}$ frame. The set of rigid camera transforms
is denoted as $\tau$, the function $\omega$ is the perspective projection, $D$ is the depth of the input frame, and $I$ is the gradient of the luminance of frame's color.

\subsection{Magnetic Localization System}
\begin{figure}[thb]
\centering
\includegraphics[width=0.8\columnwidth]{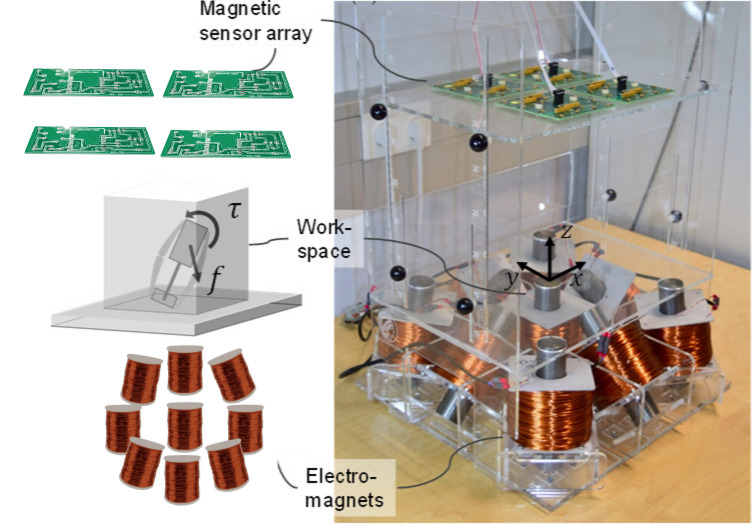}
\caption{Magnetic Localization System}
\label{fig:magnetic_sensor} 
\end{figure}

Our magnetic localization technique \cite{son20165} is able to measure 5-DoF absolute pose values for untethered meso-scale magnetic robot. As described in Figure \ref{fig:magnetic_sensor}, the system consists of a magnetic sensor system for localization and electromagnets for actuation of the magnetic capsule robot. A Hall-effect sensor array measures magnetic field at several locations from the magnetic capsule robot, whereas a computer-controlled electromagnetic coil array provides actuator's magnetic field. The core idea of our localization technique is separation of capsule's magnetic field from actuator's known magnetic field, which is realized by subtracting actuator's magnetic field component from the measured magnetic data. Finally, noise effects are reduced by second-order directional differentiation. For curious readers, further details of the magnetic localization technique can be found in \cite{son20165}.

\subsection{Deep learning based Sensor Fusion}
\begin{figure}
\centering
\includegraphics[width=0.8\columnwidth]{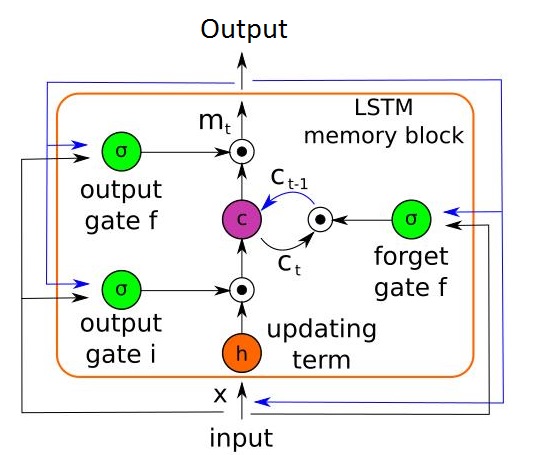}
\caption{Information flow through the hidden units of the LSTM.}
\label{fig:lstm} 
\end{figure}

Recurrent Neural Networks (RNNs) are very suitable for modelling the dependencies across data sequences and for creating a temporal motion model thanks to its memory of hidden states over time. This allows the pose estimation for the current time to benefit from the prior information of past sensor data in a similar way to the way that statistical filters use prior distributions to estimate posterior ones. To address the vanishing and exploding gradients that is the most common challenge in designing and training RNNs, a particular form of RNN, which is called Long Short-Term Memory (LSTM), was introduced by \cite{hochreiter1997long}. The LSTM model has a memory cell $c_t$ that encodes the knowledge that is observed up to time step $t$. \textit{Gates} control the behaviour of the cell. They are the layers that are multiplicatively applied, and can keep or discard a value from the gated layer. Three gates are used in the LSTM, which control whether to forget the current cell value in the forget gate, if it should read its input in the input gate and whether to output the new cell value in the output gate. The LSTM updates according to the following equations:

\begin{align}
i_t &= \sigma(W_{ix}x_t + W_{ih}h_{t-1}) \\
f_t &= \sigma(W_{fx}x_t + W_{fh}h_{t-1}) \\
g_t &= tanh(W_{gx}x_t + W_{gh}h_{t-1}) \\
c_t &= f_k\odot c_{t-1} + i_t\odot g_t \\
o_t &= \sigma(W_{ox}x_t + W_{oh}h_{t-1}) \\
h_t &= o_t\odot tanh(c_t)
\end{align}
where $\odot$ is the element-wise multiplication with a gate value, $\sigma(\cdot)$ is the sigmoid non-linearity,  $f$ is the forget gate, $i$ is the input gate, $o$ is the output gate, $g$ is the input modulation gate, and $W$ terms denote the entries of corresponding weight matrices (see Figure \ref{fig:lstm} for the visual reference).

Our deep RNN model is constructed by concatenating the core-LSTM on top of two multi-rate LSTMs with inputs from the EVO and the magnetic localization system as illustrated in Fig. \ref{fig:deep_architecture}. Each LSTM layer has $200$ hidden states. The system learns translational and rotational movements simultaneously. To regress the 6-DoF pose, we trained the RNN architecture on Euclidean loss using the Adam optimization method with the following objective loss function:
\begin{equation} \label{eqn:loss}
loss(I) = \|\hat{\mathbf{x}} - \mathbf{x}\|_2 + \beta\|\hat{\mathbf{q}} - \mathbf{q}\|_2
\end{equation}
where $\mathbf{x}$ is the translation vector and $\mathbf{q}$ is the Euler vector for a rotation. A balance $\beta$ must be kept between the orientation and translation loss values which are highly coupled as they are learned from the same model weights \cite{kendall2015posenet}. Experimental results showed that the optimal $\beta$ was given by the ratio between expected error of position and orientation at the end of training session.

The back-propagation algorithm is used to determine the gradients of RNN weights. These gradients are passed into the Adam optimization method which is a stochastic gradient descent algorithm based on estimation of first and second-order moments. The moments of the gradient are calculated using exponential moving average in addition to exponentially decaying average of past gradients, which also corrects the bias. The parameters are updated at each iteration according to the following equations: 
\begin{equation} \label{eqn:weight_update1}
(m_t)_i = \beta_1 (m_{t-1})_i + (1-\beta_1 )(\nabla L(W_t))_{i}
\end{equation}
\begin{equation} \label{eqn:weight_update2}
(v_t)_i = \beta_2 (v_{t-1})_i + (1-\beta_2)(\nabla L(W_t))_i^2
\end{equation}
\begin{equation} \label{eqn:weight_update3}
(W_{t+1})_i = (W_t)_i - \alpha \frac{\sqrt{1-(\beta_2)_i^t}}{1- (\beta_1)_i^t} \frac{(m_t)_i}{\sqrt{(v_t)_i+\varepsilon}}
\end{equation}
where the default values of the parameters $\beta_1,\beta_2$, and $\varepsilon$ are used \cite{kingma2014adam}: $\beta_1=0.9$, $\beta_2=0.999$ and $\varepsilon=10^{-8}$. 

\section{Experimental Setup}
\label{sec:exp_setup}

\subsection{Magnetically Actuated Soft Capsule Endoscopes}

\begin{figure}[b]
\centering
\includegraphics[scale=0.65]{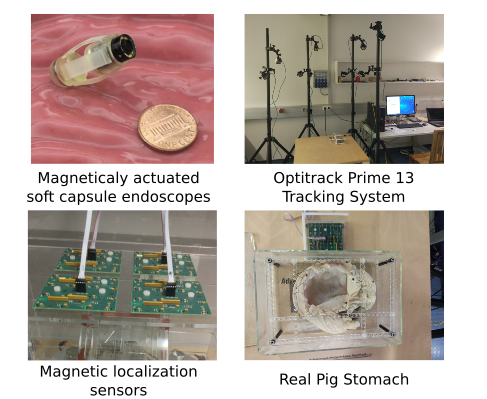}
\caption{An overview of experimental setup}
\label{fig:expset} 
\end{figure}

Our capsule prototype is a magnetically actuated soft capsule endoscope (MASCE) which is designed to be used in the upper gastrointestinal tract for disease detection, drug delivery, and biopsy operations. The prototype is composed of an RGB camera, a permanent magnet, and a drug chamber (see Fig. \ref{fig:expset} for visual reference). The magnet produces force and torque in response to a controlled external magnetic field, which are used to actuate the capsule robot and to release drugs in specifically targeted locations. A 2D magnetic sensor array is placed on top of the workspace and electromagnets are placed at the bottom of the workspace, with the patient located in between. Magnetic fields from the electromagnets generate the magnetic force and torque on the ring magnet around MASCE so that the robot moves inside the workspace. The coordinate system in Fig.\ref{fig:magnetic_sensor} shows the origin and orientation of the workspace.

\subsection{Dataset} 
\label{sec:dataset}

\begin{figure}
\centering
\includegraphics[scale=0.30]{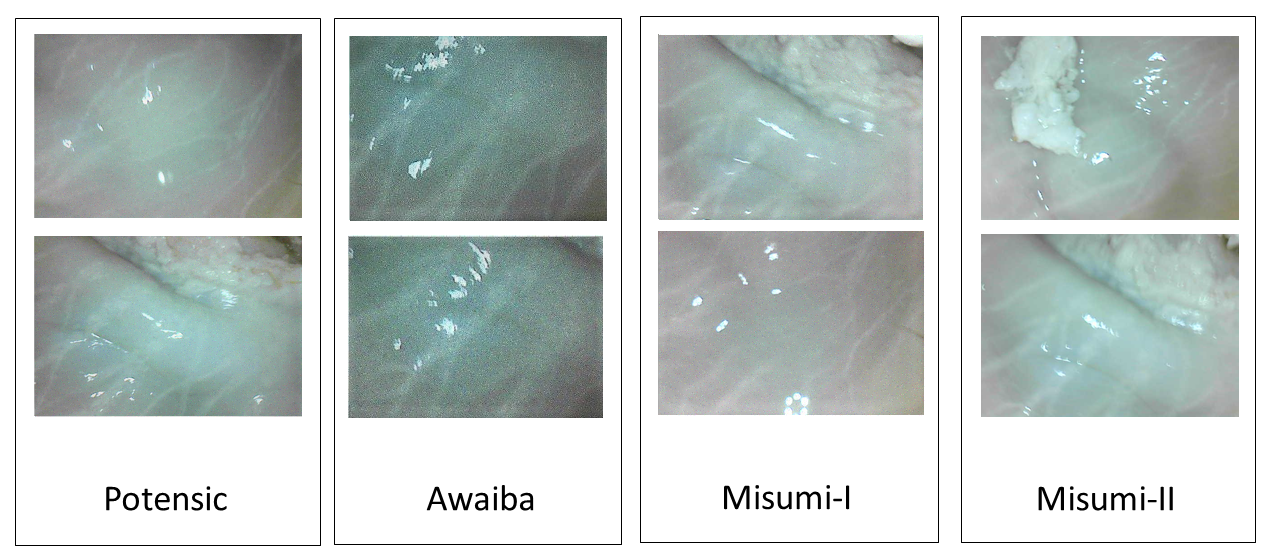}
\caption{Sample images from real pig stomach dataset}
\label{fig:dataset} 
\end{figure}

This section introduces the experimental setup and explains how the  training and testing datasets were created. The dataset was recorded on five different real pig stomachs (see Fig.\ref{fig:expset}). In order to ensure that our algorithm is not tuned to a specific camera model, four different commercial endoscopic cameras were employed. For each pig stomach and camera combination, $3000$ frames were acquired,  which makes $60000$ frames for four cameras and five pig stomachs in total.  $40000$ frames were used for training the RNNs, whereas the remaining $20000$ frames were used for evaluation. Sample real pig stomach frames are shown in Fig. \ref{fig:dataset} for visual reference. During video recording, an Optitrack motion tracking system consisting of eight Prime-13 cameras and the manufacturer's tracking software was utilized to obtain 6-DoF localization ground-truth-data with sub-millimetre accuracy (see Fig. \ref{fig:expset}) which was used as a gold standard for the evaluations of the pose estimation accuracy. The tracking system consistently produces positional error less than $0.3$ mm and rotational error less than $0.05^{\circ}$.

\section{Results and Discussion}
\label{sec:results}

\begin{figure*}[t!]
\centering
	\begin{subfigure}[t]{0.45\textwidth}  
		\includegraphics[width=\textwidth]{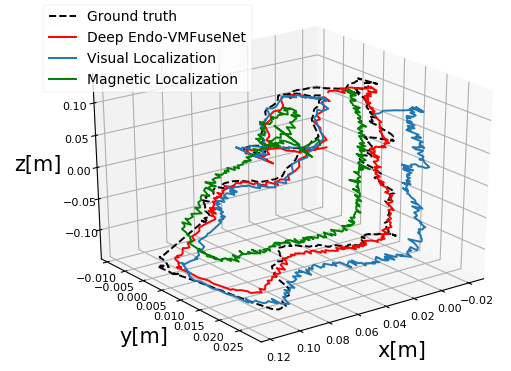}
		\caption{Trajectory 1}
		\label{fig:traj_1}
	\end{subfigure}
	\begin{subfigure}[t]{0.45\textwidth} 
\includegraphics[width=\textwidth]{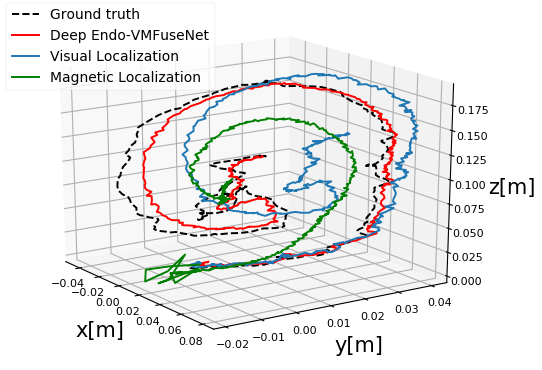}
		\caption{Trajectory 2}
		\label{fig:traj_2} 
	\end{subfigure}
	\begin{subfigure}[t]{0.45\textwidth} 
		\includegraphics[width=\textwidth]{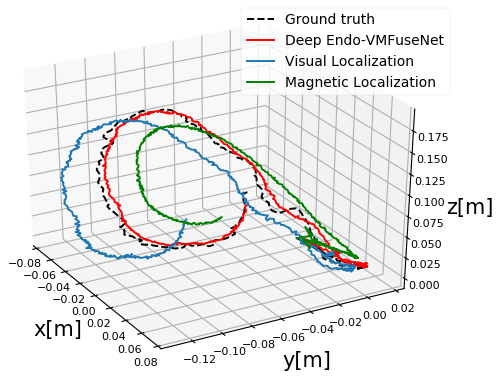}
		\caption{Trajectory 3}
		\label{fig:traj_3} 
	\end{subfigure}
	\begin{subfigure}[t]{0.45\textwidth} 
		\includegraphics[width=\textwidth]{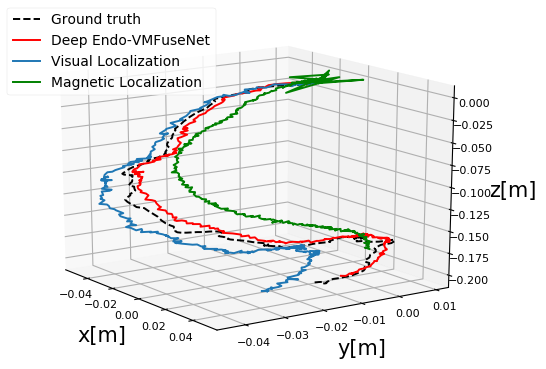}
		\caption{Trajectory 4}
		\label{fig:traj_4} 
	\end{subfigure}
	\caption{Sample ground-truth trajectories and estimated trajectories predicted by the DL-based sensor fusion approach. As seen, deep Endo-VMFuseNet is the closest to the ground truth trajectories.}
	\label{fig:trajectories}
\end{figure*}
\begin{figure*}[b!]
	\begin{subfigure}[t]{0.5\textwidth} 
		\includegraphics[width=\textwidth]{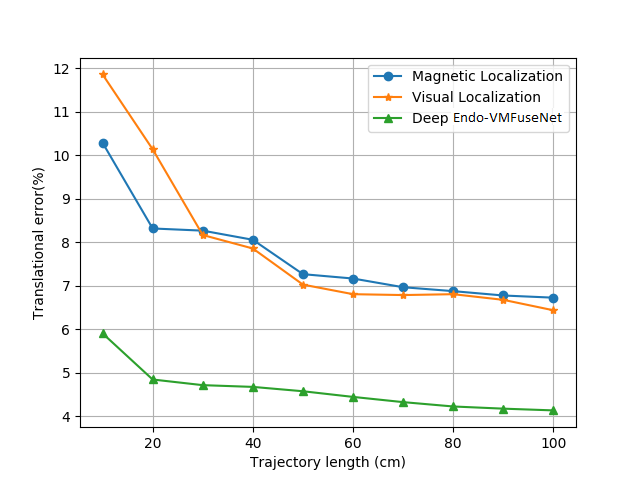}
		\caption{Trajectory length vs translation error}
		\label{fig:trans_error} 
	\end{subfigure}
	\begin{subfigure}[t]{0.5\textwidth} 
		\includegraphics[width=\textwidth]{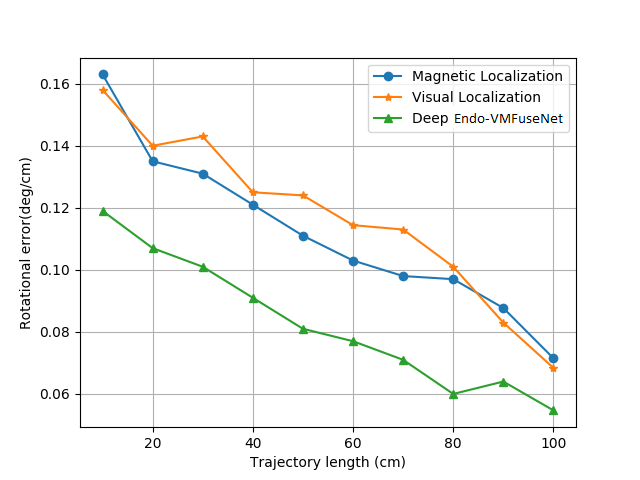}
		\caption{Trajectory length vs rotation error}
		\label{fig:rot_error}       
	\end{subfigure}
	\caption{Deep Endo-VMFuseNet outperforms both of the other models in terms of translational and rotational position estimation.}
	\label{fig:error_vs_length}
\end{figure*}

The RNN architecture was trained using the Caffe library on an NVIDIA Tesla K80 GPU. Using the back-propagation-through-time method, the weights of the hidden units were trained for up to $200$ epochs with an initial learning rate of $0.001$. Overfitting,which would make the resulting pose estimator inapplicable in other scenarios, was prevented using dropout and early stopping techniques. The dropout regularization technique, which samples a part of the whole network and updates its parameters based on the input data \cite{srivastava2014dropout}, is an extremely effective and simple method to avoid overfitting. Early stopping is another widely used technique to prevent overfitting of a complex neural network architecture optimized by a gradient-based method. We strictly avoided the use of any image frames from the training session for the testing session. The performance of the deep Endo-VMFuseNet approach was analysed using averaged Root Mean Square Error (RMSE) estimation for translational and rotational motions. For trajectories of various complexity, such as uncomplicated paths with slow incremental translations and rotations, and comprehensive scans with many local loop closures and complex paths with fast rotational and translational movements, we performed  tests on deep Endo-VMFuseNet comparing with EVO localization and magnetic localization. The average translational and rotational RMSEs for deep Endo-VMFuseNet, EVO localization and magnetic localization against different path lengths are shown  in Fig. \ref{fig:error_vs_length}, respectively. The results indicate that deep Endo-VMFuseNet clearly outperforms both EVO alone and magnetic localization alone. We presume that the effective use of the LSTM architecture in Endo-VMFuseNet architecture enabled learning from asynchronous and uncalibrated sensor array. The results also indicate that Endo-VMFuseNet is capable of handling asynchronous data (50 Hz magnetic data and 30 Hz visual data), by interpreting the localization information from the current magnetic data and previous visual and magnetic localization information saved by internal hidden memory of LSTM units. Moreover, we can conclude that Endo-VMFuseNet is also able to handle asymmetric sensor data, i.e the missing 6th degree of magnetic localization by making use of existing 6-DoF EVO and 5-DoF magnetic sensor information from current and previous frames. Some sample ground-truth and estimated trajectories for Endo-VMFuseNet, EVO localization and magnetic localization are shown in Fig.\ref{fig:traj_4} for visual reference. As seen in sample trajectories, Endo-VMFuseNet is able to stay close to the ground-truth pose values for even complex, fast rotational and translational motions, where both EVO and magnetic localization by themselves clearly deviate from the ground-truth trajectory. Thus, we can conclude that Endo-VMFuseNet makes effective use of both sensor data streams.  

\section{CONCLUSIONS}
In this study, we presented, to the best of our knowledge, the first sensor fusion method based on deep learning for endoscopic capsule robots. The proposed fusion architecture is able to achieve simultaneous learning and sequential modelling of motion dynamics across sensor streams by concatenating the core LSTM with two multi-rate LSTMs. Many issues faced by traditional sensor fusion techniques such as external calibration of sensors, synchronization between sensors and issue of missing degrees of freedoms are successfully handled by deep Endo-VMFuseNet. Since it is trained in an end-to-end manner, there is no need to carefully hand-tune the parameters of the system. In the future, we will incorporate controlled actuation into the scenario to investigate a more complete system, and additionally we will seek ways to make the system robust against representational singularities in the rotation data.

\FloatBarrier

\bibliographystyle{IEEEtran}
\balance

\bibliography{bibfile}

\end{document}